\documentclass{article}

    \usepackage[final, nonatbib]{neurips_2025}

\usepackage[T1]{fontenc}    %
\usepackage{hyperref}       %
\usepackage{url}            %
\usepackage{booktabs}       %
\usepackage{amsfonts}       %
\usepackage{nicefrac}       %
\usepackage{microtype}      %
\usepackage[dvipsnames]{xcolor}
\usepackage{graphicx}
\usepackage{colortbl}
\usepackage{multirow}
\usepackage{amsmath}
\usepackage[maxnames=100]{biblatex}
\definecolor{realtimecolor}{rgb}{1.,1,0.85}
\definecolor{CheckGreen}{rgb}{0, 0.55, 0}
\definecolor{XRed}{RGB}{180,0,0}
\newcommand{\xmark}{{\color{XRed}\ding{55}}}
\usepackage{eucal}
\usepackage{amssymb}%
\usepackage{pifont}%
\newcommand{\cmark}{{\color{CheckGreen}\ding{51}}}%

\newtoggle{comments}
\toggletrue{comments}
\iftoggle{comments}{%
    \newcommand{\tarasha}[1]{{\leavevmode\color{magenta}[Tarasha: #1]}}
    \newcommand{\deva}[1]{{\leavevmode\color{blue}[Deva: #1]}}
    \newcommand{\kaihua}[1]{{\leavevmode\color{blue}[Kaihua: #1]}}
}{%
  \newcommand{\tarasha}[1]{}
  \newcommand{\deva}[1]{}
  \newcommand{\kaihua}[1]{}
}

\definecolor{tabfirst}{rgb}{1, 0.7, 0.7}
\definecolor{tabsecond}{rgb}{1, 0.85, 0.7}
\definecolor{tabthird}{rgb}{1, 1, 0.7}
\definecolor{tabgray}{rgb}{0.9, 0.9, 0.9}

\title{Reconstruct, Inpaint, Test-Time Finetune: \\Dynamic Novel-view Synthesis from Monocular Videos}

\author{%
  Kaihua Chen\thanks{Equal contribution.} \quad
  Tarasha Khurana\footnotemark[1] \quad
  Deva Ramanan \\
  Carnegie Mellon University\\
\url{https://cog-nvs.github.io/}
}

\begin{document}

\maketitle

\begin{figure}[h]
    \centering
    \includegraphics[width=1.0\linewidth]{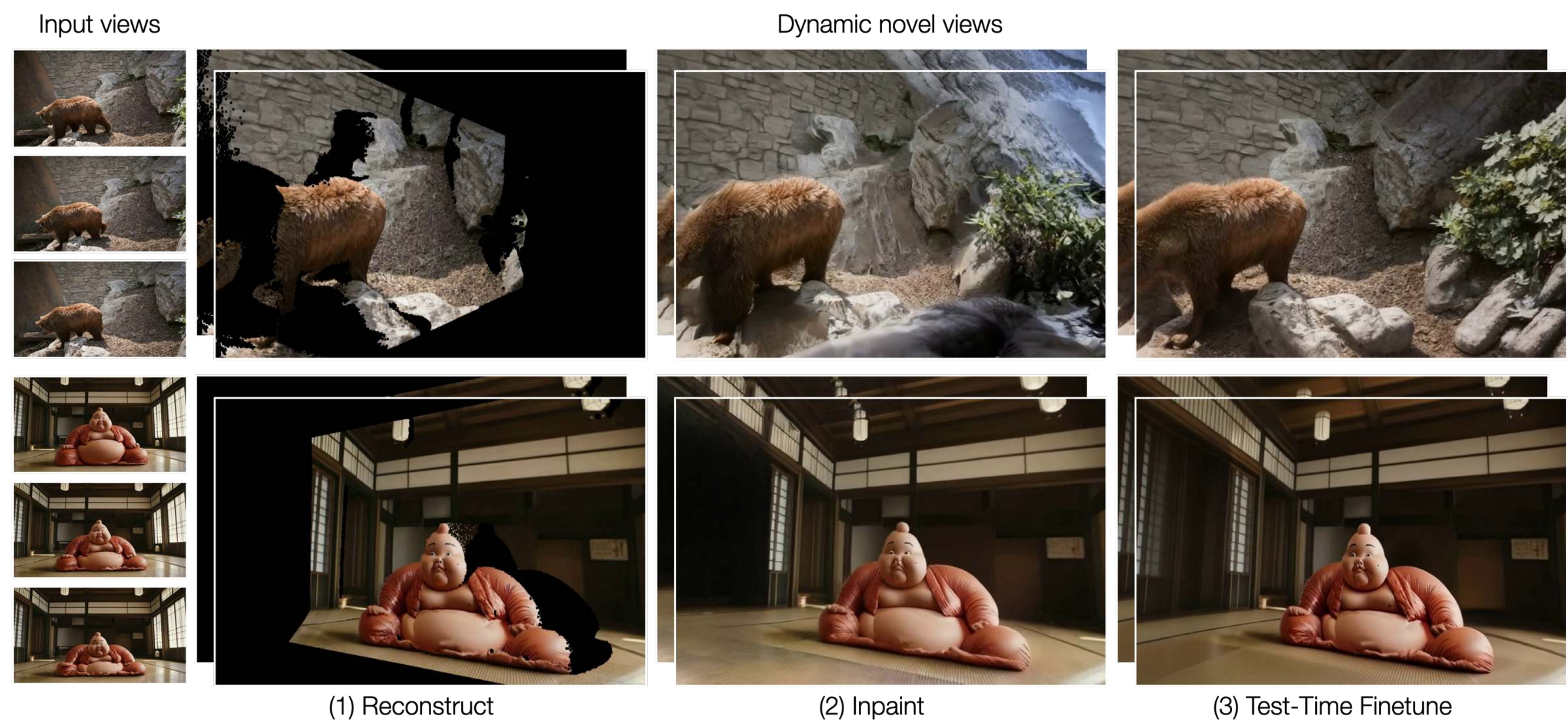}
    \caption{We present CogNVS, a video diffusion model that enables novel-view synthesis of dynamic scenes. Given an in-the-wild monocular video of a dynamic scene, we first reconstruct the scene, render it from the target novel-view and inpaint any unobserved regions. Because CogNVS can be pre-trained via self-supervision, it can also be test-time-finetuned on a given target video, enabling it to zero-shot generalize to novel domains. Our simple pipeline outperforms almost all prior state-of-the-art for dynamic novel-view synthesis. We show outputs from CogNVS from two unseen videos; a real-world video above, and a generated video below.}
    \label{fig:splash}
\end{figure}

\begin{abstract}
  We explore novel-view synthesis for dynamic scenes from monocular videos. Prior approaches rely on costly test-time optimization of 4D representations or do not preserve scene geometry when trained in a feed-forward manner. Our approach is based on three key insights: (1) {\em covisible} pixels (that are visible in both the input and target views) can be rendered by first reconstructing the dynamic 3D scene and rendering the reconstruction from the novel-views and (2) {\em hidden} pixels in novel views can be ``inpainted" with feed-forward 2D video diffusion models. Notably, our video inpainting diffusion model (CogNVS) can be self-supervised from 2D videos, allowing us to train it on a large corpus of in-the-wild videos. This in turn allows for (3) CogNVS to be applied zero-shot to novel test videos via {\em test-time finetuning}. We empirically verify that CogNVS outperforms almost all prior art for novel-view synthesis of dynamic scenes from monocular videos.
\end{abstract}

\section{Introduction}

Rapid advances in static 3D scene representations \cite{mildenhall2020nerf, kerbl20233dgs} have paved the way for spacetime understanding of the dynamic world. This has enabled photorealistic content creation and immersive virtual reality applications. In this work, we focus on the problem of novel-view synthesis from casually-captured monocular videos of dynamic scenes. 

\paragraph{Why is this hard?} Prior work on dynamic view synthesis addresses this task from two extremes. The first class of methods ``test-time'' optimize a new 4D representation from scratch for every new test video. While this ensures physically-plausible scene geometry, careful choices in modeling scene motion -- in the form of an independent deformation field, or learnable temporal offsets -- have to be made~\cite{wang2024shapeom,DynIBaR, 4dgswu}. %
More importantly, it can take on the order of hours to optimize and render a  novel-view video. An attractive alternative is to train large feed-forward video models {\em directly} for view synthesis~\cite{zhang2024gslrm,honglrm}. While inference on such models is dramatically faster (on the order of milliseconds), the resulting renderings often are not as accurate as their test-time optimized counterparts. From a pragmatic perspective, such models need to be trained on mega-scale multi-view training data, which is difficult to obtain for dynamic scenes.

\paragraph{Our method} addresses the above challenges by decomposing the problem of dynamic view-synthesis into three distinct stages. First, we lean on the success of non-rigid structure from motion \cite{lu2024align3r, zhang2024monst3r, li2024megasam} approaches that produce reconstructions of visible scene regions, sometimes known as "2.5D" reconstructions (since occluded regions are not reconstructed). We point out that such reconstructions can be trivially produced for casual mobile videos captured with depth sensors and egomotion~\cite{iphone}. 
When such reconstructions are rendered from a target novel view, previously-hidden regions will not not be rendered. To ``inpaint'' these regions, we train a 2D video-inpainter -- CogNVS -- by fine-tuning a video diffusion model (CogVideoX \cite{yang2024cogvideox}) to condition on the partially-observable novel-view pixels. Importantly, we allow CogNVS to {\em also} update the appearance of previously-visible pixels, allowing our pipeline to model view-dependent (dynamic) scene effects. %

\vspace{7pt}
The \textbf{key insight} of our work is that CogNVS can be trained on any 2D video via self-supervision.  However, rather than training our inpainter with random 2D masks, we make use of 3D multi-view supervision that better captures 3D scene visiblity, similar to prior art~\cite{weber2024nerfiller}. %
Specifically, given a 2D training video, we first reconstruct it (with an off-the-shelf method such as MegaSAM) and then render the reconstruction from a random camera trajectory. This rendering is used to identify co-visible pixels from the source video that remain visible in the novel views. This original source video and its co-visible-only masked variant can now form a training pair for 3D-consistent video inpainting. %
Importantly, because such a training pair does {\em not} require ground-truth 3D supervision, CogNVS can be trained on diverse in-the-wild 2D videos. We use dynamic scenes from TAO \cite{dave2020tao}, SA-V \cite{ravi2024sam}, Youtube-VOS \cite{xu2018youtube}, and DAVIS \cite{davis}. Equally as important, we use the same paradigm to {\em test-time finetune} CogNVS on the test video-of-interest. We show that this allows our pipeline to ``zero-shot'' generalize to test videos that were never seen during training. We argue that our test-time finetuning of 2D diffusion models can be seen as the ``best-of-both-worlds'', by leveraging large-scale training data (for data-driven robustness) and test-time optimization (for accuracy).

\vspace{5pt}
In summary, our contributions are as follows: (1) We decompose dynamic view synthesis into three stages of reconstruction, inpainting and test-time finetuning, (2) we use a large corpus of \textit{only 2D videos} for training CogNVS, and (3) we do extensive \textit{zero-shot} benchmarking on three evaluation datasets against state-of-the-art methods and show improvements on dynamic view synthesis.

\section{Related Work}

\paragraph{Novel-view synthesis} has seen recent advancements with the rise of implicit scene representations like NeRFs \cite{mildenhall2020nerf} and Gaussian primitives \cite{kerbl20233dgs, keselman2022fuzzy, chen2024mvsplat}. We have seen widespread efforts in scaling these representations to model larger scenes \cite{turki2022mega, xiangli2021citynerf, tancik2022block}, making them faster to fit \cite{lee2025fast, garbin2021fastnerf, fan2024instantsplat, kerbl20233dgs, muller2022instant, cao2023hexplane, deng2022depth, chen2021mvsnerf}, anti-aliased \cite{liang2025analytic, barron2021mip, barron2022mip, hu2023Tri-MipRF, barron2023zip}, and extend to representing dynamic scenes \cite{stearns2024marbles, gao2024gaussianflow, pumarola2021d, lee2025fast}. The most popular paradigms have been the adoption of dynamic NeRFs \cite{mildenhall2020nerf, pumarola2021d, DynNeRF, nerfies} and deformable Gaussian primitives \cite{kerbl20233dgs, 4dgswu, yang2023gs4d, luiten2024dynamic} for modeling scene dynamics, apart from using voxel grids \cite{khurana2023point, khurana2022differentiable} or learnable tokenzation \cite{zhang2023copilot4d}. Most approaches need multi-view posed videos as input, and only recently monocular view synthesis has gained traction \cite{gafni2021dynamic, lei2024mosca, wang2024shapeom, li2021neural}. However, each of the aforementioned approaches have to be test-time optimized separately for every new test video, are slow to optimize and yet fail to recover highly-detailed dynamic scene content \cite{lei2024mosca}. Moreover, there is no focus on predicting the unobservable scene content, which is exacerbated by benchmarking metrics that only evaluate co-visible pixels \cite{iphone} in training and inference views and therefore encourage benchmarking on novel views that are not too far apart from the training views. Our approach instead reformulates dynamic view synthesis as an inpainting task, which specifically focuses on generating parts of the scene that were occluded from the training views, thereby facilitating extreme novel view synthesis for dynamic scenes. Our large-scale pretraining for feed-forward novel-view inpainting enables data-driven robustness.

\begin{figure}[t]
    \centering
    \includegraphics[width=0.95\linewidth]{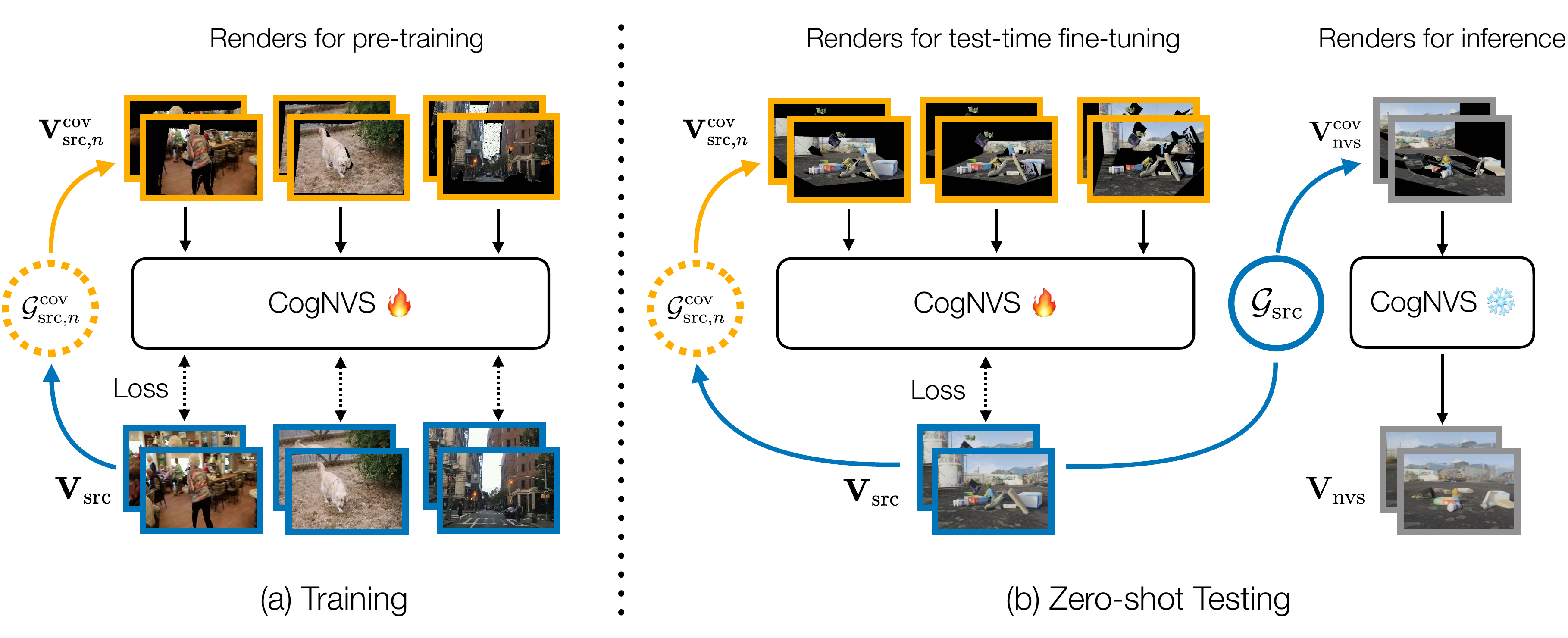}
    \caption{\textbf{CogNVS overview.} During training (\textbf{left}), given a 2D source video (in \textcolor{blue}{blue}) of a dynamic scene, we first reconstruct the scene using off-the-shelf monocular reconstruction algorithms like MegaSAM \cite{li2024megasam} to obtain the 3D scene geometry, $\mathcal{G}_{\rm src}$ and camera odometry, $\mathbf{c}_{\rm src}$. We then sample a set of arbitrary camera trajectories $\{\mathbf{c}_1, \cdots, \mathbf{c}_N\}$ to simulate plausible occluded geometries, $\{\mathcal{G}^{\rm cov}_{{\rm src},1}, \cdots, \mathcal{G}^{\rm cov}_{{\rm src},N}\}$ which when rendered from original camera trajectory, $\mathbf{c}_{\rm src}$ produces a mask of source pixels that are co-visible in the sampled trajectory (in \textcolor{orange}{orange}). The source video and its masked variant produce a self-supervised training pair for learning CogNVS, our video inpainting diffusion model (visualized in Fig.~\ref{fig:datageneration}). At inference (\textbf{right}), we finetune CogNVS on the given input sequence by similarly constructing self-supervised training pairs. The final novel-view is then generated using the finetuned CogNVS in a feed-forward manner.}
    \label{fig:method}
\end{figure}

\paragraph{Data-driven novel-view synthesis} approaches have emerged \cite{gcd, wu2024cat4d, ren2025gen3c, yu2025trajectorycrafter, yu2024viewcrafter, liu2023zero, yang2024storm} which train for view synthesis in a feed-forward manner with large-scale data. One class of methods is based on transformer architectures, more often than not trained with multi-view supervision and rendering in the loop \cite{honglrm, zhang2024gslrm, jiang2024real3d, wei2024meshlrm, xie2024lrm, ren2024l4gm}. Another class of methods reformulate novel-view synthesis as a conditional generation task and use diffusion-based generative architectures \cite{liu2023zero, shi2023zero123++} for the same. Initially, the focus was on developing data-driven pipelines for static novel-view synthesis \cite{liu2023zero, liu2023one, liu2024one, yu2024viewcrafter, zhang2024gslrm, wei2024meshlrm, chan2023genvs, 3dim}, or exploiting data-driven priors \cite{poole2022dreamfusion, wang2023prolificdreamer, sun2023dreamcraft3d, chen2024using, Tang_2023_makeit, khurana2024predicting, raj2023dreambooth3d, you2024solver}, using multi-view posed image inputs. However, the focus is now shifting to dynamic view synthesis of casually captured videos in an unconstrained setting \cite{gcd, ren2025gen3c, yu2025trajectorycrafter, yang2024storm, li2024spacetime, jeong2024dreammotion}. These approaches allow a greater level of hallucination of unseen scene components which instills the capability of view synthesis for camera poses that are far apart from the training views. We fit into this setting. While the data-driven learning provides faster inference times and a broad generalization, it compromises on 3D geometric accuracy and physical-plausibility which introduces unrealistic artifacts in the synthesized outputs (\textit{e.g.}, objects suddenly exist or cease to exist) \cite{yu2025trajectorycrafter}. In this work, we highlight that test-time finetuning is crucial to preserving the 3D geometry of the scene and reducing implausible artifacts.

\paragraph{Test-time finetuning} is a long-standing paradigm to curb distribution shifts in machine learning algorithms and improve their generalization. It's origin lies in early-age algorithms for optical character recognition \cite{bottou1992local} and text classification \cite{joachims1999transductive}, where the algorithm adjusts itself \textit{after} observing the test data. A decade back, it popularly resurfaced for super-resolution \cite{shocher2018superres} where learning to super-resolve an image was achieved by downsampling and super-resolving the test image. Domain generalization approaches for vision \cite{sun2020ttt, chen2022ttt, iwasawa2021ttt, chen2023ttt, gandelsman2022maskedautoencoders, zhao2024mfh} soon took inspiration from this breakthrough and recently, chain-of-thought prompting \cite{wei2022cot} and general LLM reasoning \cite{akyurek2024fewshot} in natural language processing adapted this paradigm. The most recent adoption was seen in 4D reconstruction and tracking \cite{st4rtrack2025}, and we similarly explore this paradigm further in our work.

\section{Method}

Given a monocular video of a dynamic scene, $\mathbf{V}_{\rm src} = \{\mathbf V_{\rm src}^{t}\}_{t=1}^{T}$, we want to generate a novel view of the observed scene, $\mathbf{V}_{\rm nvs} = \{\mathbf V_{\rm nvs}^{t}\}_{t=1}^{T}$ from a target camera pose. As discussed (c.f. Fig. \ref{fig:method}), we achieve this by decomposing the task into three distinct stages -- (1) obtain an off-the-shelf reconstruction of the observed scene over time, (2) render the scene from the novel views and inpaint the non-co-visible regions, and (3) curb the train-test distribution shift with test-time finetuning. Note that the first two stages of our pipeline are similar to concurrent work \cite{yu2025trajectorycrafter, ren2025gen3c} but importantly we find that test-time finetuning is a crucial stage to allow generalization. We now describe each of the stages in detail.

\subsection{Dynamic view synthesis as structured inpainting}
We use off-the-shelf SLAM frameworks, like MegaSAM \cite{li2024megasam}, to obtain a reconstruction of the given scene. Formally, let the underlying 3D structure of the world as observed by $\mathbf{V}_{\rm src}$ be represented by, $\mathcal{G}_{\rm src} = \{\mathbf X_{\rm src}^t\}_{t=1}^T$, where $\mathbf X_{\rm src}^t$ are the evolving 3D primitives (points, Gaussians, etc.) across time, $t$. Any physical properties of the primitives are omitted from this discussion for simplicity. Let the recovered camera poses from which $\mathbf{V}_{\rm src}$ was observed be, $\mathbf{c}_{\rm src} = \{\mathbf c_{\rm src}^t\}_{t=1}^T$, where $\mathbf c$ denotes a camera pose and is formulated as,  \(\mathbf c=(\mathbf R,\mathbf t)\in\mathrm{SE}(3)\) lie group. The source video $\mathbf{V}_{\rm src}$ can be obtained by using a rendering function $\mathcal{R}$ as,
\[
    \mathbf{V}_{\rm src} = \mathcal{R}\bigl(\mathcal{G}_{\rm src}, \mathbf{c}_{\rm src}\bigr)
\]

\paragraph{Learning to inpaint novel views} For obtaining $\mathbf{V}_{\rm nvs}$, we note that a subset of 3D primitives that must be visible from $\mathbf{c}_{\rm src}$, are already available in the reconstructed scene geometry, $\mathcal{G}_{\rm src}$. Therefore, a partial observation of the world in the form of \textit{co-visible} pixels \cite{iphone} from novel views, $\mathbf{c}_{\rm nvs} = \{c_{\rm nvs}^t\}_{t=1}^T$, can be rendered as follows,
\[
    \mathbf{V}_{\rm nvs}^{\rm cov} = \mathcal{R}\bigl(\mathcal{G}_{\rm src}, \mathbf{c}_{\rm nvs}\bigr)
\]

At this point, the novel view synthesis is incomplete, and all missing regions have to be generated. To this end, we train a conditional video diffusion model, CogNVS (denoted by $\epsilon_\theta$) built on top of a recently proposed transformer-based video diffusion model \cite{yang2024cogvideox}. CogNVS takes in the partially observed novel view video and generates an inpainted novel-view of the scene. 
The overall CogNVS pipeline first employs a 3D causal VAE to compress the conditioning $\mathbf{V}_{\rm nvs}^{\rm cov}$ and target novel-view $\mathbf{V}_{\rm src}$ into latent representations $\mathbf z_{\rm cond} = \mathcal{E}(\mathbf{V}_{\rm nvs}^{\rm cov})$ and $\mathbf z_0 = \mathcal{E}(\mathbf{V}_{\rm src})$ respectively, enabling efficient training while preserving temporal coherence and photometric fidelity. Here, $\mathcal{E}$ is the VAE encoder. Gaussian noise is then added to the target latent $\mathbf z_0$, and the resulting noisy latent is concatenated with the conditional latent $\mathbf z_{\rm cond}$. This joint representation is passed through a self-attention transformer equipped with 3D rotary positional embeddings (3D-RoPE)~\cite{su2024rope} and adaptive layer normalization, which predicts the added noise. The training objective follows a score matching formulation:
\[
\min_{\theta}\;
\mathbb{E}_{\substack{
\mathbf z_0 = \mathcal{E}(\mathbf{V}_{\rm src}),\;
\mathbf z_{\rm cond} = \mathcal{E}(\mathbf{V}_{\rm nvs}^{\rm cov}),\\
k \sim \mathcal{U}\{1,\dots,K\},\;
\epsilon \sim \mathcal{N}(0,I)}
}
\bigl\|\,
\epsilon_{\theta}(\mathbf z_k, k, \mathbf z_{\rm cond})\;-\;\epsilon
\bigr\|_2^2
\]

Here, $\mathbf z_k=\sqrt{\bar{\alpha_k}}\mathbf z_0 + \sqrt{1-\bar{\alpha_k}}\epsilon$ denotes the noisy latent at a uniformly sampled timestep $k$, where $\bar{\alpha_k}$ is the cumulative signal preserving factor. While CogVideoX was originally designed as an image-to-video diffusion model that zero-pads conditional image patches to match the length of the target video, we adapt it for a video-to-video setting, where the shapes of the conditional and target inputs are inherently aligned and no padding is needed. In practice, CogNVS is trained with datasets of 2D videos which are used to generate self-supervised training pairs. We discuss this below.

\subsection{Data generation for self-supervised training}

\begin{figure}[t]
    \centering
    \includegraphics[width=1.0\linewidth]{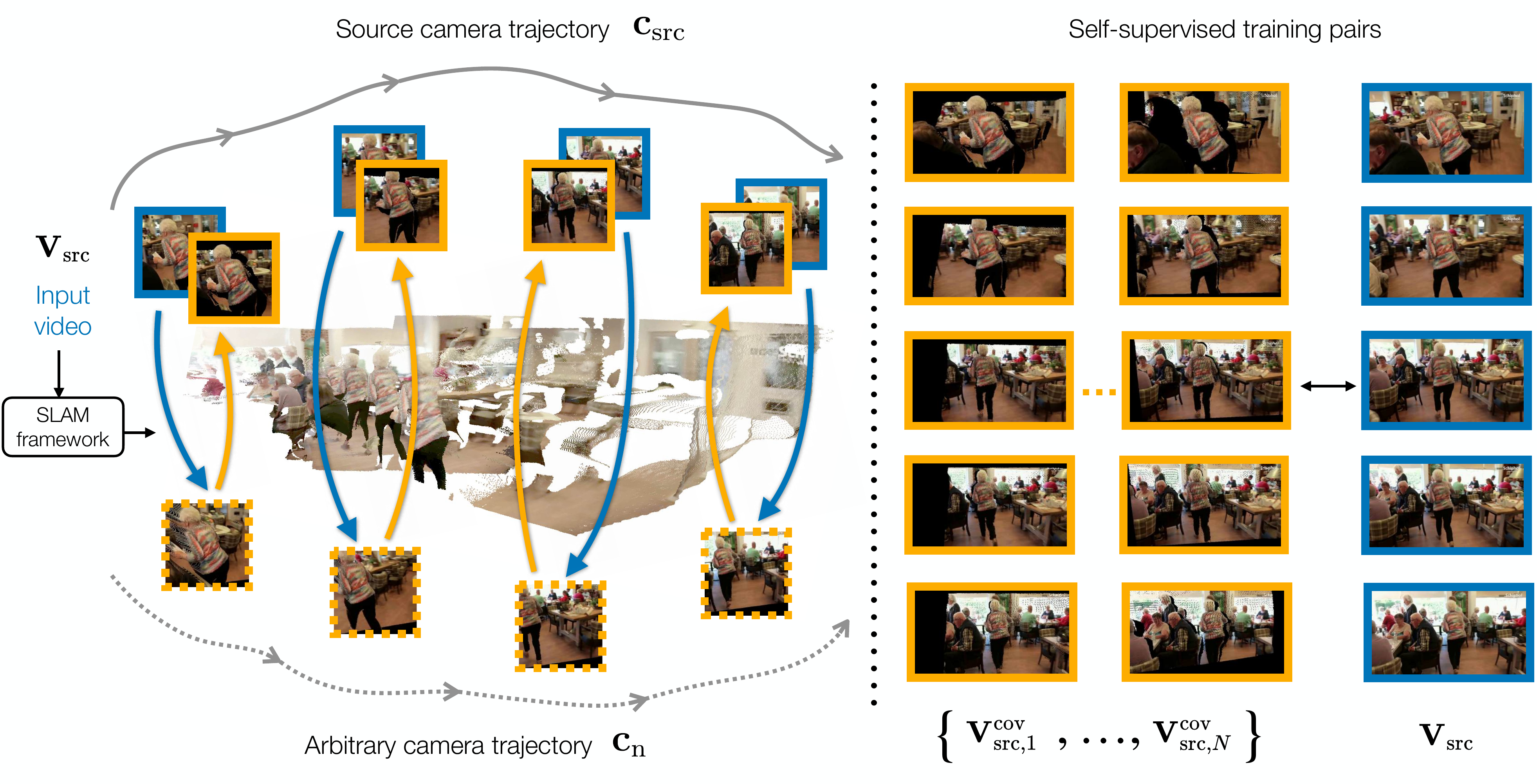}
    \caption{\textbf{Self-supervised training data generation.} To curate a large training set for video inpainting, we first reconstruct an input source 2D video (in \textcolor{blue}{blue}) with an off-the-shelf monocular SLAM system. After reconstruction, we randomly sample $N$ pairs of `start' and `end' camera poses around a spherical region, $\mathcal{S}$ of the estimated camera pose in the given 2D video. $\mathcal{S}$ is bounded by a predefined deviation in the spherical coordinate axes, similar to a prior work \cite{yu2024viewcrafter}. 
    We sample a ${\rm SE(3)}$ camera trajectory that interpolates the start and end poses while looking at the center of the scene. We render the reconstruction from this novel trajectory (in \textcolor{orange}{dotted-orange}), and use the rendering to identify co-visible pixels in the original source view (in \textcolor{orange}{orange}). The source video and its masked variant are used to produce a self-supervised training pair for training CogNVS, our "3D-aware" video inpainting diffusion model.} 
    \label{fig:datageneration}
\end{figure}

We propose to train CogNVS in a self-supervised manner. This allows us to use a large corpus of 2D videos. For each casually captured monocular video $\mathbf{V}_{\rm src}$, similar to prior work~\cite{han2022single, lv2025spatialdreamer, yu2025trajectorycrafter}, we obtain its 3D reconstruction $\mathcal{G}_{\rm src}$ and odometry $\mathbf{c}_{\rm src}$ from off-the-shelf SLAM frameworks \cite{li2024megasam}. As demonstrated in Fig. \ref{fig:datageneration}, we sample $N$ arbitrary camera trajectories in order to create training pairs from 2D videos, described as follows.

We first obtain the ``center'' of the scene by considering the pixel at the optical center in the first frame of the given video, similar to a prior work \cite{yu2025trajectorycrafter}. We then construct a bounded region $\mathcal{S}$ in spherical coordinates, around the camera center of $c_{\rm src}^1$. Within this region, we uniformly sample start and end spherical coordinates of each new camera trajectory, and then again sample two intermediate camera locations between the start and end spherical coordinates to ensure smoothness during interpolation. Camera poses are obtained by converting the spherical coordinates into euclidean space to get translations, and camera rotations are obtained such that the look-at vector always points to the center of the scene. Using the four sampled camera poses, we do bicubic interpolation on the $\rm SE(3)$ manifold. This results in a set of smooth camera trajectories, $\{\mathbf{c}_{\rm n}\}_{n=1}^{N}$ which are then used to construct the training pairs.
With $N$ trajectories, we can obtain ``partial'' novel-view renderings as,
\[
    \mathbf{V}_{n}^{\rm cov} = \mathcal{R}\bigl(\mathcal{G}_{\rm src}, \mathbf{c}_{n}\bigr)
\]

Between $\mathbf{V}_{n}^{\rm cov}$ and $\mathbf{V}_{\rm src}$, only a subset of primitives from $\mathcal{G}_{\rm src}$ are \textit{co-visible}. Let this subset be denoted by $\mathcal{G}^{\rm cov}_{{\rm src}, n}$ for the $n^{\rm th}$ trajectory. Then, partial renderings of the source video are given by,
\[
    \mathbf{V}_{{\rm src}, n}^{\rm cov} = \mathcal{R}\bigl(\mathcal{G}^{\rm cov}_{{\rm src}, n}, \mathbf{c}_{\rm src}\bigr) \hspace{0.7cm} {\rm s.t.} \hspace{0.7cm} \mathcal{D} = \{(\mathbf{V}_{{\rm src}, n}^{\rm cov}, \mathbf{V}_{\rm src})\} \forall n \in [1, N]
\]

is the set of training pairs created by one monocular video. We repeat this for all 2D videos considered.

\subsection{Test-time finetuning for target domain adaptation} At test time, to reduce domain gap arising due to different scene properties (lighting, appearance, motion) we use the source test video $\mathbf{V}_{\rm src}$ to adjust the priors of CogNVS and create self-supervised finetuning pairs, $\mathcal{D}$ as described above. We therefore adapt the model weights \(\theta\) on‐the‐fly with $M$ gradient steps with $\eta$ step size as follows,
\[
    \theta \;\leftarrow\;
    \theta \;-\;\eta\;\nabla_{\theta}
    \bigl\|\,
      \epsilon_\theta(\mathbf z_k,\,k,\,\mathbf z_{\rm cond}^n)
      -\epsilon
    \bigr\|_2^2,
\]
where $\mathbf z_{\rm cond}^n$ is the latent of the $n^{\rm th}$ self-supervised training pair input.
At the end of finetuning, we obtain the desired novel view $\mathbf V_{\rm nvs}$ from CogNVS by using the partially observed novel-view, \(\mathcal{R}(\mathcal G_{\rm src}, \mathbf c_{\rm nvs})\) as the input conditioning, and running a reverse diffusion process.

\label{sec:datagen}

\section{Empirical Analysis}

\subsection{Experimental setup}
\paragraph{Datasets} We train CogNVS on four in-the-wild video datasets, SA-V \cite{ravi2024sam}, TAO \cite{dave2020tao}, Youtube-VOS \cite{xu2018youtube}, and DAVIS \cite{perazzi2016davis}. We sample 3000, 3000, 4000 and 100 videos respectively from each of the datasets, giving us a total training video pool of $\approx$ 10,000 videos. For pretraining, we randomly select a new subsequence of 49-frames in every epoch and construct its training pairs. For benchmarking, we follow prior work \cite{lei2024mosca, gcd} and use a combination of Kubric-4D, ParallelDomain-4D \cite{gcd} and Dycheck \cite{iphone}. These have a held-out test set of 20, 20 and 5 videos each. Note that our evaluation on Kubric-4D, ParallelDomain-4D and Dycheck is zero-shot as the datasets are not seen during training. Since the Kubric-4D and ParallelDomain-4D are synthetic, we use their groundtruth point clouds and odometry for a fair comparison to baselines. For Dycheck, we use MegaSAM for reconstruction and align the estimated point cloud with the groundtruth to solve for scale ambiguity.

\paragraph{Baselines} For Kubric-4D, we consider GCD \cite{gcd} and Gen3C \cite{ren2025gen3c}, alongside a concurrent work, TrajectoryCrafter \cite{yu2025trajectorycrafter}. For ParallelDomain-4D, we consider the same baselines except Gen3C, which only evaluates on Kubric-4D, as there is no open-source implementation available yet. For Dycheck, we consider recent work like Shape-of-Motion \cite{wang2024shapeom}, MoSca \cite{lei2024mosca}, CAT4D \cite{wu2024cat4d}. Note that we do not benchmark test-time optimization approaches on Kubric-4D and ParallelDomain-4D, because their performance degrades catastrophically on novel views that are far apart from training views. For more quantitative analysis of CAT4D, see appendix.

\paragraph{Metrics} For pixel-wise photometric evaluation, we adopt the widely used PSNR, SSIM, and LPIPS family of metrics for evaluating reconstruction quality via novel-view synthesis. We additionally benchmark the generation quality with FID and KID. This is in line with the benchmarking proposed in several diffusion-based view synthesis works \cite{gcd, ren2025gen3c, liu2023zero, tung2024megascenes}. 

\paragraph{Implementation details}  During pretraining, we load the official CogVideoX-5B-I2V checkpoint and fully finetune all 42 transformer blocks. We use the AdamW optimizer with $\beta_1 = 0.9$, $\beta_2=0.95$, and $\beta_3=0.98$, a learning rate of $2\times 10e-5$, and a batch size of 8 for 12,000 steps. To fit within 48GB VRAM, we employ DeepSpeed ZeRO-2~\cite{rajbhandari2020zero} to partition model states across 8 A6000 Ada GPUs in a distributed setting. Pretraining completes in approximately 3 days. 

During test-time finetuning, we maintain the same optimizer and learning rate but reduce the number of steps to 200 for shorter sequences (e.g., Kubric-4D) and 400 for longer ones (e.g., DyCheck). For all experiments, we use an input resolution of $\mathbb{R}^{49\times480\times 720}$, set the classifier-free guidance scale to 6, and run 50 inference steps. A single novel-view sequence generates in $\sim$5 mins on an A6000 Ada. We provide additional implementation details and evaluation protocols in appendix.

\begin{figure}[t]
    \centering
    \includegraphics[width=1.0\linewidth]{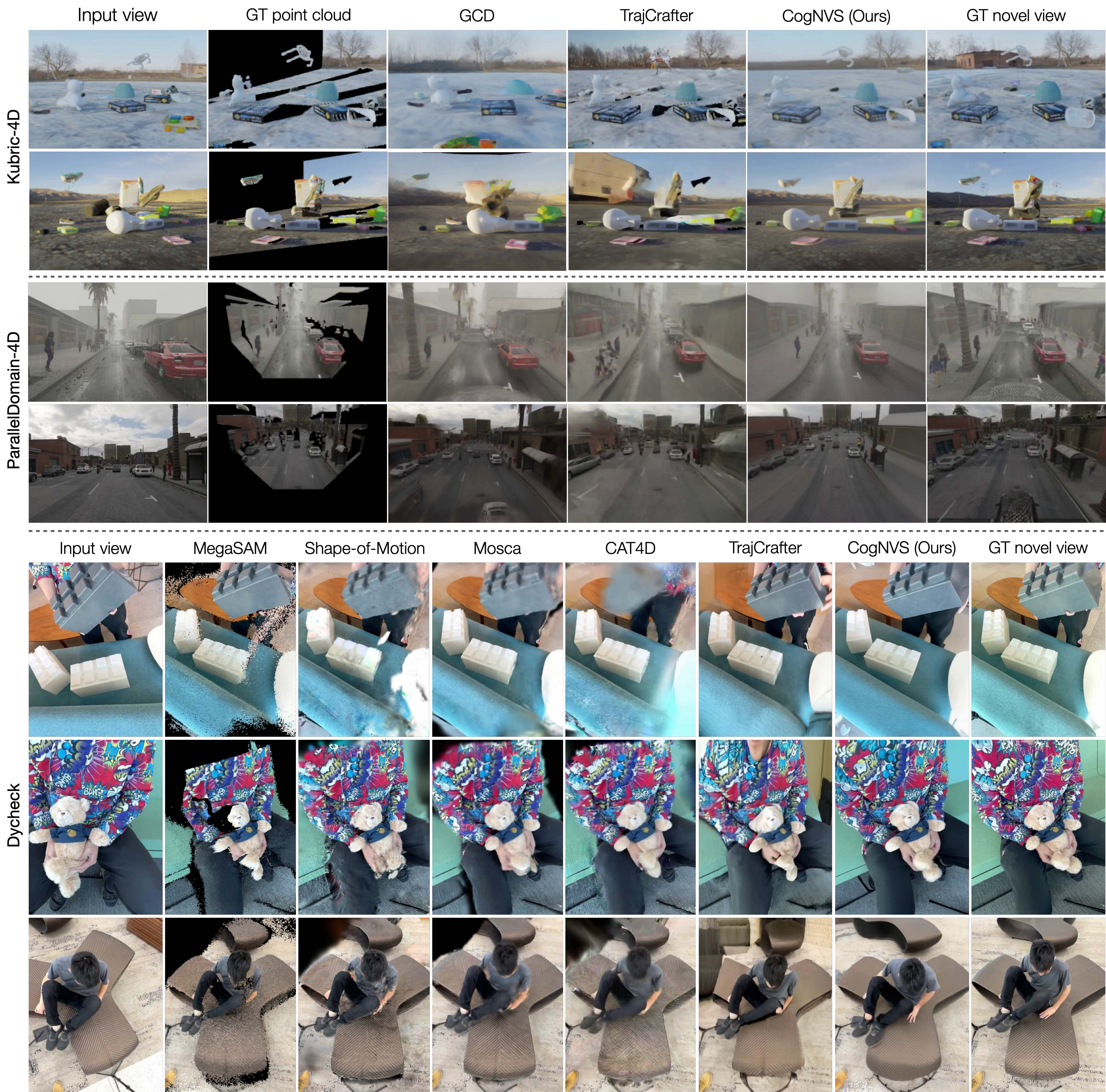}
    \caption{We show a qualitative comparison with state-of-the-art approaches for dynamic novel-view synthesis on Kubric-4D (\textbf{top}), ParallelDomain-4D (\textbf{middle}) and DyCheck (\textbf{bottom}). Note how reconstruction alone, either by groundtruth depth, MegaSAM \cite{li2024megasam}, Shape of Motion \cite{wang2024shapeom}, or MoSca \cite{lei2024mosca} cannot synthesize a complete novel view. Optimization based approaches like Shape of Motion, and MoSca, blur the dynamic regions when fitting 4D representations. CAT4D \cite{wu2024cat4d}, whose visuals are taken from its project page due to unavailable code, struggles to generalize. TrajectoryCrafter \cite{yu2025trajectorycrafter} over-hallucinates the occluded regions and does not preserve geometry. GCD \cite{gcd} performs well because it was trained on Kubric-4D and ParallelDomain-4D. Our method can instead produce photorealistic and 3D-consistent novel-views for the given scenes in a \textit{zero-shot} manner with test-time finetuning, even starting from point cloud renders that are incomplete and noisy (\textit{e.g.}, from MegaSAM for DyCheck). It is consistently able to synthesize sharp dynamic objects, which the other baselines struggle with. Please see the video in the appendix.}
    \label{fig:sota}
\end{figure}

\subsection{Comparison to state-of-the-art}

\begin{table*}[t]
\centering
\caption{Comparison to state-of-the-art for dynamic view synthesis on Kubric-4D and ParallelDomain-4D. We find that our method, that operates zero-shot unlike Gen3C and GCD, achieves state-of-the-art performance across a majority of metrics. $^\dagger$ Note that Gen3C only evaluates on Kubric-4D and there is no open-source code that would allow us to benchmark it on ParallelDomain-4D.}
\label{tab:sota}
\resizebox{\columnwidth}{!}{
\begin{tabular}{l|ccccc|ccccc}
\toprule
\multirow{2}{*}{Method} & \multicolumn{5}{c|}{Kubric-4D} & \multicolumn{5}{c}{ParallelDomain-4D} \\
& PSNR ↑ & SSIM ↑ & LPIPS ↓ & FID ↓ & KID ↓ & PSNR ↑ & SSIM ↑ & LPIPS ↓ & FID ↓ & KID ↓ \\
\midrule
GT      & 15.12 & \cellcolor{tabthird}0.671 & 0.328 & 175.01 & 0.063 & 18.79 & 0.499 & 0.409 & 197.99 & 0.129 \\
GCD \cite{gcd}   & 18.59 & 0.555 & 0.383 & \cellcolor{tabthird}121.57 & \cellcolor{tabsecond}0.020 & \cellcolor{tabsecond}21.77 & \cellcolor{tabthird}0.665 & \cellcolor{tabthird}0.400 & \cellcolor{tabfirst}90.58 & \cellcolor{tabfirst}0.022 \\
Gen3C$^\dagger$  \cite{ren2025gen3c}    & \cellcolor{tabthird}19.41 & 0.630 & \cellcolor{tabthird}0.290 & \cellcolor{tabfirst}98.58 & n/a & n/a & n/a & n/a & n/a & n/a \\
TrajCrafter \cite{yu2025trajectorycrafter}    & \cellcolor{tabsecond}20.93 & \cellcolor{tabsecond}0.730	& \cellcolor{tabsecond}0.257 & 130.20 & \cellcolor{tabthird}0.024 & \cellcolor{tabthird}21.46 & \cellcolor{tabsecond}0.719	& \cellcolor{tabsecond}0.342 & \cellcolor{tabsecond}95.38 & \cellcolor{tabsecond}0.026 \\
\midrule
CogNVS  & \cellcolor{tabfirst}22.63 & \cellcolor{tabfirst}0.760 & \cellcolor{tabfirst}0.232 & \cellcolor{tabsecond}102.47 & \cellcolor{tabfirst}0.008 & \cellcolor{tabfirst}24.34 & \cellcolor{tabfirst}0.797 & \cellcolor{tabfirst} 0.302 & \cellcolor{tabthird} 102.43 & \cellcolor{tabthird} 0.033 \\
\bottomrule
\end{tabular}}
\end{table*}

\begin{table}[t]
\centering
\caption{Comparison to state-of-the-art for dynamic novel‑view synthesis on Dycheck. First, we note that our method can be run on top of any reconstruction approach and the better the reconstruction (\textit{e.g.}, replacing MegaSAM with MoSca), the better the view synthesis. Second, we see that our method can achieve state-of-the-art FID / KID scores because test-time optimization approaches \cite{wang2024shapeom, lei2024mosca, li2024megasam} result in blurry dynamic regions and cannot hallucinate new scene content, and completely feed-forward approaches \cite{yu2025trajectorycrafter} cannot return precise geometry. Our method instead gets the ``best of both worlds''.}
\label{tab:sotadycheck}
\resizebox{0.7\columnwidth}{!}{
\begin{tabular}{lcccccccc}
\toprule
Method   & PSNR ↑ & SSIM ↑ & LPIPS ↓ & FID ↓ & KID ↓ \\
\midrule
MegaSAM \cite{li2024megasam}       & 12.16 & 0.299 & 0.698 & 239.57 & 0.148 \\
Shape-of-Motion \cite{wang2024shapeom}      & \cellcolor{tabthird}15.30 & \cellcolor{tabfirst}0.476 & \cellcolor{tabfirst}0.494 & 164.29 & 0.073 \\
MoSca \cite{lei2024mosca}    & \cellcolor{tabsecond}16.22 & \cellcolor{tabsecond}0.472 & \cellcolor{tabsecond}0.586 & 148.18 & 0.063 \\
TrajCrafter \cite{yu2025trajectorycrafter} & 12.74 & 0.337 & 0.749 & \cellcolor{tabthird} 140.35 & \cellcolor{tabthird} 0.059 \\
\midrule
CogNVS (MegaSAM) & 	15.19 &	0.382 & 0.622 & \cellcolor{tabsecond} 94.48 &	\cellcolor{tabfirst}0.030\\
CogNVS (MoSca) &  \cellcolor{tabfirst}16.94 & \cellcolor{tabthird}0.449 & \cellcolor{tabthird} 0.598	& \cellcolor{tabfirst}92.83 & \cellcolor{tabsecond}0.031  \\
\bottomrule
\end{tabular}}
\end{table}

\paragraph{Kubric-4D and ParallelDomain-4D} We first do zero-shot benchmarking of CogNVS on two synthetic datasets that come with high-fidelity dense depth and accurate camera odometry annotations. For a fair comparison to all baselines, we use the depth and poses to backproject the given scene into a canonical coordinate frame. Given this scene, we generate self-supervised pairs for test-time finetuning. Upon inference (see Tab. \ref{tab:sota}), we find that CogNVS beats prior work on photometric evaluation with PSNR, SSIM, LPIPS, even when baselines are not evaluated zero-shot (GCD is trained on Kubric-4D and ParallelDomain-4D and Gen3C is trained on Kubric-4D). In Fig. \ref{fig:sota}, we show the plausible and realistic novel-views predicted by our method on both datasets as compared to the baselines. This is quantitatively demonstrated by better FID and KID scores. A concurrent work, TrajectoryCrafter \cite{xiao2024trajectory} performs competitively. We also evaluate the rendered visible scene structure from groundtruth depth, for establishing a lower bound on dynamic-view synthesis. 

\paragraph{DyCheck} We evaluate the performance of our method on a real-world dataset of casually captured iPhone videos in Tab. \ref{tab:sotadycheck}. First, note that since CogNVS can be applied on top of reconstructions from any method, we show two variants. Better initial reconstruction (in this case, with MoSca rather than MegaSAM) allows for better dynamic view synthesis. Second, of all approaches, our method produces the most visually plausible novel views, as captured by drastically better FID and KID. Third, note how TrajectoryCrafter \cite{yu2025trajectorycrafter}, also based on video diffusion which was the second-best method on Kubric-4D, is unable to handle the distribution shift in Dycheck (shallow field-of-view, close-up videos of moving objects) and fails to generalize. Whereas our method benefits from test-time finetuning and is able to adjust to any new data-distribution at test-time. Other test-time optimization approaches (Shape-of-Motion, MoSca) do better as long as evaluation views are close to training views, because there is only one distribution they need to fit to.

\paragraph{Runtime analysis} We conduct a runtime analysis of CogNVS against other state-of-the-art baselines, as shown in Tab. \ref{tab:runtime}, covering both stages of test-time optimization and rendering/inference. Specifically, we measure runtime by evaluating each method on the DyCheck evaluation set, with an average video length of 400 frames. Under our default hardware configuration, CogNVS requires 140 min for 400 steps of test-time fine-tuning on these long videos, followed by an additional 5 minutes for rendering. Note that this duration can be flexibly traded off with computational cost, which has recently been referred to as ``inference-time scaling'' in diffusion models~\cite{ma2025inference}. For example, as shown in Fig. \ref{fig:ablation_curves} (left) in the appendix, our method achieves 96\% of its final performance with only half the fine-tuning steps.

\begin{table}[t]
\centering
\caption{Runtime analysis of test-time optimization and feed-forward approaches on DyCheck. We report the time taken for optimization / test-time finetuning for all approaches in addition to the final inference / rendering duration. In general, our optimization is faster than some test-time optimization approaches with the additional benefit of being able to inpaint/hallucinate unknown regions in a spatiotemporally consistent manner. Additionally, our inference is on par with other feed-forward methods.}
\label{tab:runtime}
\resizebox{0.7\columnwidth}{!}{
\begin{tabular}{lcccccccc}
\toprule
Method   & Optimization & Rendering / Inference \\
\midrule
MegaSAM \cite{li2024megasam} & 9 min & Real-time \\ 
MoSca \cite{lei2024mosca} & 66 min & Real-time \\
Shape of Motion \cite{wang2024shapeom} & 237 min & Real-time \\
GCD \cite{gcd} & - & 2 min \\
TrajCrafter \cite{yu2025trajectorycrafter}  & - & 5 min \\
CogNVS & 140 min & 5 min \\
\bottomrule
\end{tabular}}
\end{table}

\subsection{Ablation studies}

\begin{figure}[t]
    \centering
    \includegraphics[width=0.7\linewidth]{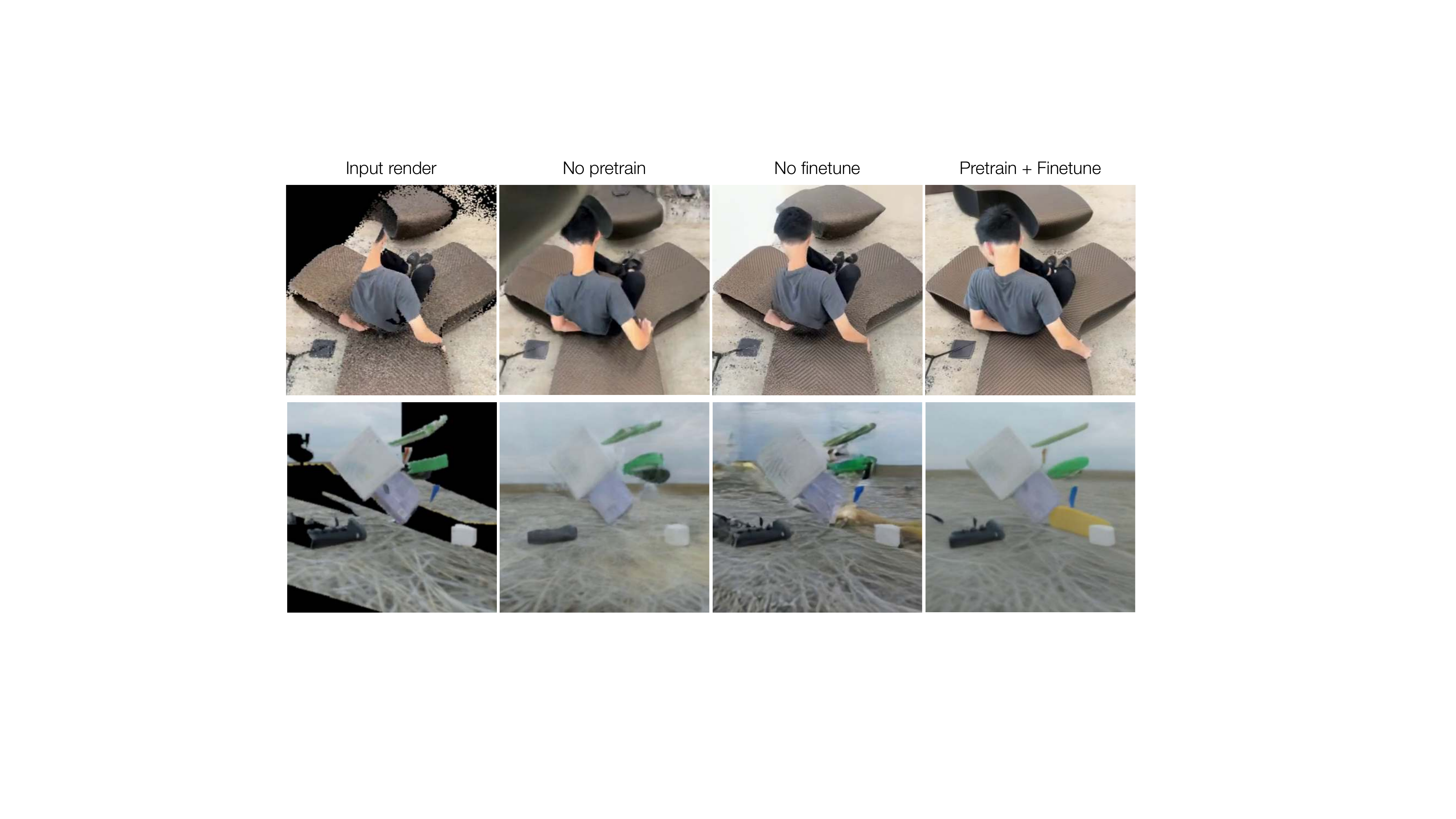}
    \caption{We qualitatively analyze the effect of pretraining and test-time finetuning. We note that without the data-driven robustness and generalization of pretraining (\textbf{second column}), CogNVS cannot hallucinate missing regions properly (\textit{e.g.}, inpainted region in first row is still black in top left corner). Finally, without test-time finetuning (\textbf{third column}), 3D consistency and adherence to scene lighting and appearance properties cannot be ensured (\textit{e.g.}, overall darker scene in second row, and output off by a few pixels at the bottom and right side of the image in first row, thereby inhibiting geometric consistency).}
    \label{fig:ablation}
\end{figure}

\paragraph{Effect of test-time finetuning}
We study the effectiveness of the test-time finetuning stage of our method. Row 2 vs. 3 in Tab. \ref{tab:ablation} show that proposed self-supervised finetuning is crucial for adaptation of CogNVS to a target video's distribution at test-time. Once the self-supervised test-time finetuning stage is completed, our method yields outputs with high fidelity, showcasing improved precision, and more contextually and geometrically consistent 3D appearances, as shown in Fig. \ref{fig:ablation}.

\begin{table}[t]
\centering
\caption{We ablate our design choices of large-scale pretraining and test-time finetuning on three randomly chosen sequences from Kubric-4D test set. We find that no pretraining is detrimental to the performance of CogNVS, so much so that the PSNR drops by 5 points, thereby devoiding CogNVS of data-driven robustness. Test-time finetuning is also essential as without the adaptation of CogNVS to the test video, the performance in terms of PSNR drops by $\sim$ 3 points.}
\label{tab:ablation}
\resizebox{0.7\columnwidth}{!}{
\begin{tabular}{cccccccccc}
\toprule
Pretrain & Finetune & PSNR ↑ & SSIM ↑ & LPIPS ↓ & FID ↓ & KID ↓ \\
\midrule
\xmark & \cmark & 18.62 & 0.691 & 0.318 & 201.29 & 0.051 \\
\cmark & \xmark & 20.06 & 0.662 & 0.284 & 185.48 & 0.038\\
\cmark & \cmark & \textbf{23.29} & \textbf{0.779} & \textbf{0.240} & \textbf{158.98} & \textbf{0.036}  \\
\bottomrule
\end{tabular}}
\end{table}

\paragraph{Effect of large-scale pretraining}
We also study the usefulness of the large-scale pretraining stage with 2D videos from 4 training datasets. In this case, test-time finetuning alone with a self-supervised objective, cannot pull CogNVS out of the local minima it reaches without a good initialization. This is a common failure mode of many test-time optimization approaches that overfit to the training views but default to rendering artifacts such as blurry dynamic regions \cite{lei2024mosca}. We show in Tab. \ref{tab:ablation} (Row 1 vs. 3) and Fig. \ref{fig:ablation} that pretraining is essential for data-driven robustness.

\paragraph{Effect of reconstruction quality}
Although we touch upon how the initial reconstruction affects the quality of dynamic view synthesis, we describe in detail here. We create a pertubed version of the Kubric-4D dataset, by obtaining reconstruction and odometry from MegaSAM and aligning the reconstruction to groundtruth to solve for scale ambiguity. Quantitative results show a $\sim$ 3 points drop on PSNR and a consistently worse performance on all metrics with sub-optimal reconstructions and cameras. This also addresses the gap in the photometric performance (with PSNR, SSIM, LPIPS) of MegaSAM-based CogNVS on DyCheck. For the quantitative and qualitative analysis of this ablation, please see the appendix.

\section{Discussion}

In this work, we focus on the problem of dynamic novel-view synthesis from monocular videos. Contrary to prior state-of-the-art that approaches this task from two extremes (either test-time optimization for every new video from scratch, or large-scale feed-forward novel view synthesis) -- we propose a simple setup that is the ``best-of-both-worlds". We reformulate dynamic view synthesis as an inpainting task and lean on the success of reconstruction algorithms like MegaSAM that can estimate the structure and geometry of in-the-wild videos. We first train a video inpainter, CogNVS, on pairs of co-visible novel-view pixels and target novel-views via self-supervision on only 2D videos. At test-time, we propose to finetune CogNVS, again via self-supervision, to adjust to the target video distribution. The proposed setup provides data-driven robustness with the large-scale pretraining of a video inpainting model, and enhances 3D accuracy of the predictions with test-time finetuning.

\paragraph{Limitations} CogNVS does not currently take advantage of open-source 3D and 4D video datasets and trains on a relatively small set of 2D videos. While the zero-shot evaluation can achieve better photorealistic performance than prior state-of-the-art even with this unprivileged training data, the model and its geometric inpainting capabilities can be enhanced by adding more training data from all three -- 2D, 3D and 4D data sources. Additionally, the performance of CogNVS is dependent on the quality of dynamic scene reconstruction obtained from off-the-shelf structure from motion algorithms. When groundtruth structure and odometry is available, such as from ubiquitous depth sensors, CogNVS's performance can be increased. A limitation of the data generation pipeline is that the sampled arbitrary camera trajectories are not able to mimic the diversity of camera trajectories that are encountered in real-life, which is a bottleneck to the performance of CogNVS. A better strategy would be to create a ``data-driven'' trajectory sampler that samples from a set of real-world trajectories observed in the training set.

{\bf Acknowledgements}: This work was supported in part by Bosch and by the Intelligence
Advanced Research Projects Activity (IARPA) via Department of Interior/Interior Business Center (DOI/IBC) contract
number 140D0423C0074. The U.S. Government is authorized to reproduce and distribute reprints for Governmental
purposes notwithstanding any copyright annotation thereon. Disclaimer: The views and conclusions contained herein are those of the authors and should not be interpreted as necessarily representing the official policies or endorsements, either expressed or implied, of IARPA, DOI/IBC, or the U.S. Government.

\printbibliography

\appendix

\newpage
\begin{center}

{\bf \Large Appendix}
    
\end{center}

\vspace{10pt}

In this appendix, we extend our discussion of dynamic view synthesis in casual monocular videos. First, we discuss the intricacies in the training and evaluation protocols adopted. This is followed by an in-depth ablation study on various design decisions in the proposed pipeline. Finally, we show more qualitative results, both on the considered benchmarks and on in-the-wild examples.

\section{Implementation Details}

\paragraph{Training pair details} 
To generate self-supervised training pairs, we randomly perturb the source camera trajectory to create diverse camera paths. In the spherical coordinate system, we sample random elevations from 
[-15$^\circ$, 15$^\circ$], azimuths from [-30$^\circ$, 30$^\circ$], and radius deviations from [-0.15, 0.15], followed by bicubic interpolation. This procedure enables flexible generation of training pairs across arbitrary camera trajectories. For pretraining, we use $N=2$ camera views per training videos. For test-time finetuning, we set $N=5$ for DyCheck and $N=9$ for both Kubric-4D and ParallelDomain-4D, due to their wider novel-view gaps. When a video sequence exceeds CogVideoX's default input length of 49 frames, we randomly sample a 49-frame subsequence in each epoch. On DyCheck, we additionally apply a noise injection strategy to simulate real-world degradation on training pairs, as analyzed in Section B.

\paragraph{Evaluation protocol} 
For Kubric-4D and ParallelDomain-4D, we follow the official GCD evaluation protocol, using the 20-sequence sub–test set and an input resolution of 576 (width) × 384 (height). On DyCheck, we follow the official Shape-of-Motion and Mosca evaluation protocols, and we consistently report evaluation metrics at an image resolution of 720 (width) × 960 (height) on the five standard test sequences: apple, block, paper-windmill, spin, and teddy. We render dynamic Gaussian representations from Mosca and Shape-of-Motion with a black background for fair comparison. Both methods optimize camera poses using the ground-truth novel views to improve photometric metrics; we retain this step to stay consistent with their original implementation. CAT4D, although diffusion-based, fits a 4D-GS representation (with minor extensions) after synthesizing multi-view videos. When evaluating CogNVS on MegaSAM renders, we append a static background extracted from the full input video to better capture long-term context. The effectiveness of background stacking is validated in Section B. Also note that Shape-of-Motion and MoSca optimize for evaluation camera poses during evaluation using ground-truth novel view videos. Whether CAT4D adopts this step is unknown. We do not do this camera pose optimization at test-time.
 Since the DyCheck evaluation sequences are more than 49-frames in length, we isolate the static scene regions and stack them in 3D across time. This accumulated background is then rendered onto each frame which helps, to a large extent, ``pre-inpaint'' the static background regions using fused information from multiple 49-frame length sequences.

\begin{table}[h]
\centering
\caption{Effect of reconstruction quality on Kubric-4D. We quantitatively evaluate CogNVS's performance with the use of two different reconstructions for Kubric-4D. Groundtruth depth gives an upperbound on view synthesis performance by CogNVS. Our first observation, perhaps unsurprisingly, is that the quality of MegaSAM reconstruction is subpar to that of the groundtruth. This difference is quality is also translated to the novel-view synthesis task with CogNVS, where CogNVS used with groundtruth depth does 3 and 45 points better at PSNR and FID respectively as compared to CogNVS used on top of MegaSAM.}
\label{tab:recon_quality}
\resizebox{0.7\columnwidth}{!}{
\begin{tabular}{cccccccccc}
\toprule
Method & PSNR ↑ & SSIM ↑ & LPIPS ↓ & FID ↓ & KID ↓ \\
\midrule
MegaSAM & 12.73 & 0.299 & 0.644 & 280.62 & 0.164 \\
GT & 15.12 & 0.671 & 0.328 & 175.01 & 0.063 \\
CogNVS (MegaSAM) & 19.62 & 0.621 & 0.313 & 147.83 & 0.033 \\
CogNVS (GT) & \textbf{22.63} & \textbf{0.760} & \textbf{0.232} & \textbf{102.47} & \textbf{0.008}\\
\bottomrule
\end{tabular}}
\end{table}

\begin{figure}[h]
    \centering
    \includegraphics[width=\linewidth]{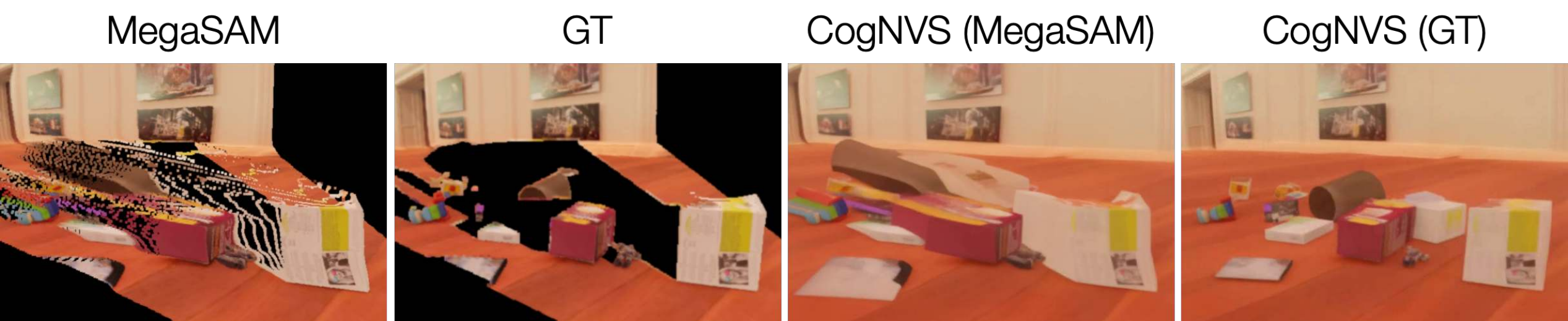}
    \caption{We show the effect of using different qualities of reconstruction. Note that the groundtruth depth of the scene is perfect because it is derived synthetically. This re-rendered depth results in more realistic object placements in the scene as compared to the predictions using the depth from MegaSAM. This is because the MegaSAM depth is noisy at the object edges and therefore results in smeared objects in the novel view predictions.}
    \label{fig:ablation_noisy_kubric}
\end{figure}

\begin{table}[h]
\centering
\caption{Effect of masking strategy on Kubric-4D. We study the effect of building CogNVS as an inpainting model using other masking strategies, specifically, random and tube masking \cite{tong2022videomae}. We find that random masking is the least optimal as it does not mimic the test-time scenario, tube masking does better, but our structured masking strategy is the best for the proposed structured inpainting task.}
\label{tab:masking}
\resizebox{0.6\columnwidth}{!}{
\begin{tabular}{cccccccccc}
\toprule
Mask & PSNR ↑ & SSIM ↑ & LPIPS ↓ & FID ↓ & KID ↓ \\
\midrule
Random & 20.62 & 0.755 & 0.310 & 187.79 & 0.059 \\
Tube & 21.75 & 0.778 & \textbf{0.236} & 173.55 & 0.041 \\
Ours & \textbf{23.29} & \textbf{0.779} & 0.240 & \textbf{158.98} & \textbf{0.036}\\
\bottomrule
\end{tabular}}
\end{table}

\begin{figure}[h]
    \centering
    \includegraphics[width=\linewidth]{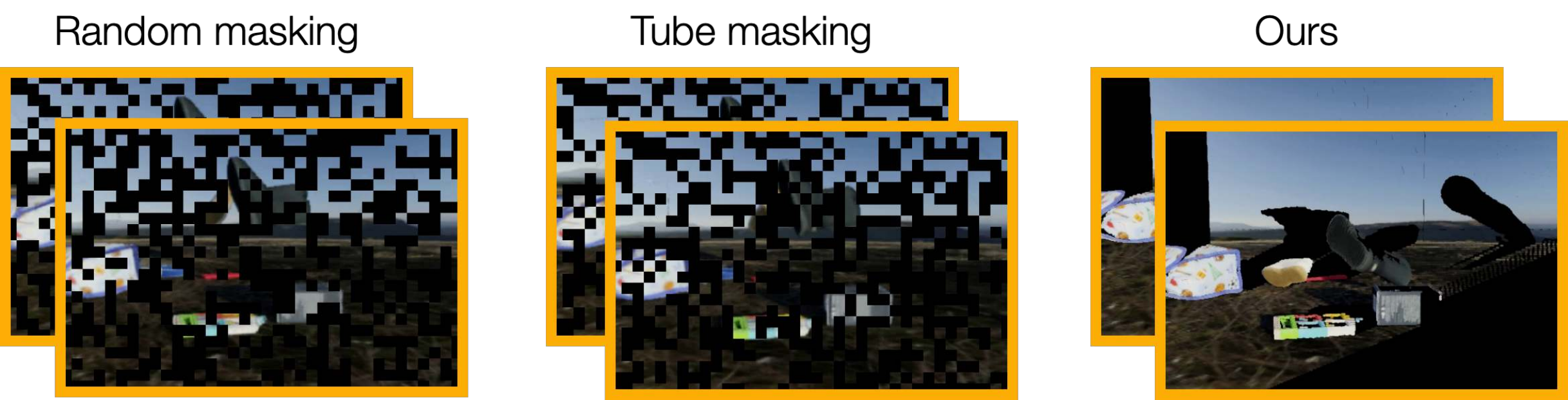}
    \caption{We illustrate the different masking strategies considered, as proposed by a prior work \cite{tong2022videomae}. For random masking (\textbf{left}), the masked out patches are different in each frame of the input video. For tube masking (\textbf{center}), a random set of patches is masked but this set is constant across multiple frames of the video. For our structured masking (\textbf{right}), we derive the mask by rendering visible scene reconstruction from the novel views.}
    \label{fig:ablation_masks}
\end{figure}

\section{Ablation Study}

\paragraph{Effect of reconstruction quality} In Tab. \ref{tab:recon_quality}, we show the effect of using different sources of reconstructions on the entire Kubric-4D evaluation set.  Specifically, we compare the structure and odometry from MegaSAM \cite{li2024megasam} and the synthetic depth groundtruth from Kubric-4D \cite{gcd}. We find that both quantitatively and qualitatively (c.f. Fig. \ref{fig:ablation_noisy_kubric}), our pipeline benefits more from better reconstructions. This is because the quality of reconstruction directly affects the input to CogNVS, and if the input point cloud is noisy (\textit{e.g.}, smearing at the object borders), the prediction of the novel view also becomes inaccurate.

\paragraph{Ablation on masking strategy} Since CogNVS is an inpainting model, we ablate different masking strategies to train CogNVS on three sequences from Kubric-4D, instead of the proposed structured masking. In Tab. \ref{tab:masking} and Fig. \ref{fig:ablation_masks}, we use random and tube masking from VideoMAE \cite{tong2022videomae} and apply them with a 50\% masking ratio on the input video sequences divided into 16 $\times$ 16 patches. We find that random masking is the least optimal as it does not resemble the structured inpainting task at test-time. Tube masking is more amenable to the test-time inpainting pattern, which reflects as better photometric and generative metrics. Of all, our structured masking obtained by rendering scene reconstructions into the novel views performs the best.

\begin{figure}[h]
    \centering
    \includegraphics[width=\linewidth]{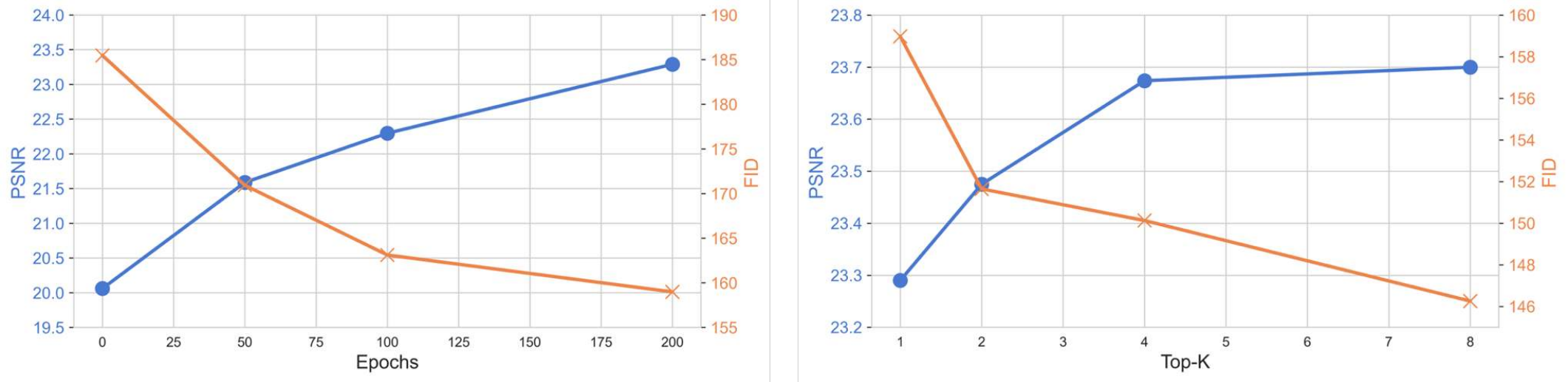}
    \caption{We conduct ablations on the number of epochs used for test-time finetuning (\textbf{left}) and number of samples drawn from CogNVS for a probabilistic evaluation (right). Both experiments suggest similar trends; performance improves with an increase in the number of finetuning epochs and increase in the number of samples drawn from our diffusion model. Performance saturates once a threshold is reached.}
    \label{fig:ablation_curves}
\end{figure}

\paragraph{Ablation on test-time finetuning epochs} Following the same data setup as above, we assess how the length of test-time finetuning affects the final prediction from CogNVS. In Fig. \ref{fig:ablation_curves} (left), as expected we observe that the performance improvement in the first few epochs is high (both in terms of PSNR and FID going from 0 to 50 epochs) and saturates as the number of epochs are increased further (up to 200).

\paragraph{Top-K evaluation} Following the same data setup as above, we compute probabilistic PSNR and FID metrics for CogNVS's performance on Kubric-4D in the form of Top-k metrics (where the best of k number is reported) in Fig. \ref{fig:ablation_curves} (right). As we sample multiple modes from CogNVS's learnt distribution, the Top-k metrics for PSNR and FID become better and start to saturate near k=8.

\begin{figure}[h]
    \centering
    \includegraphics[width=0.8\linewidth]{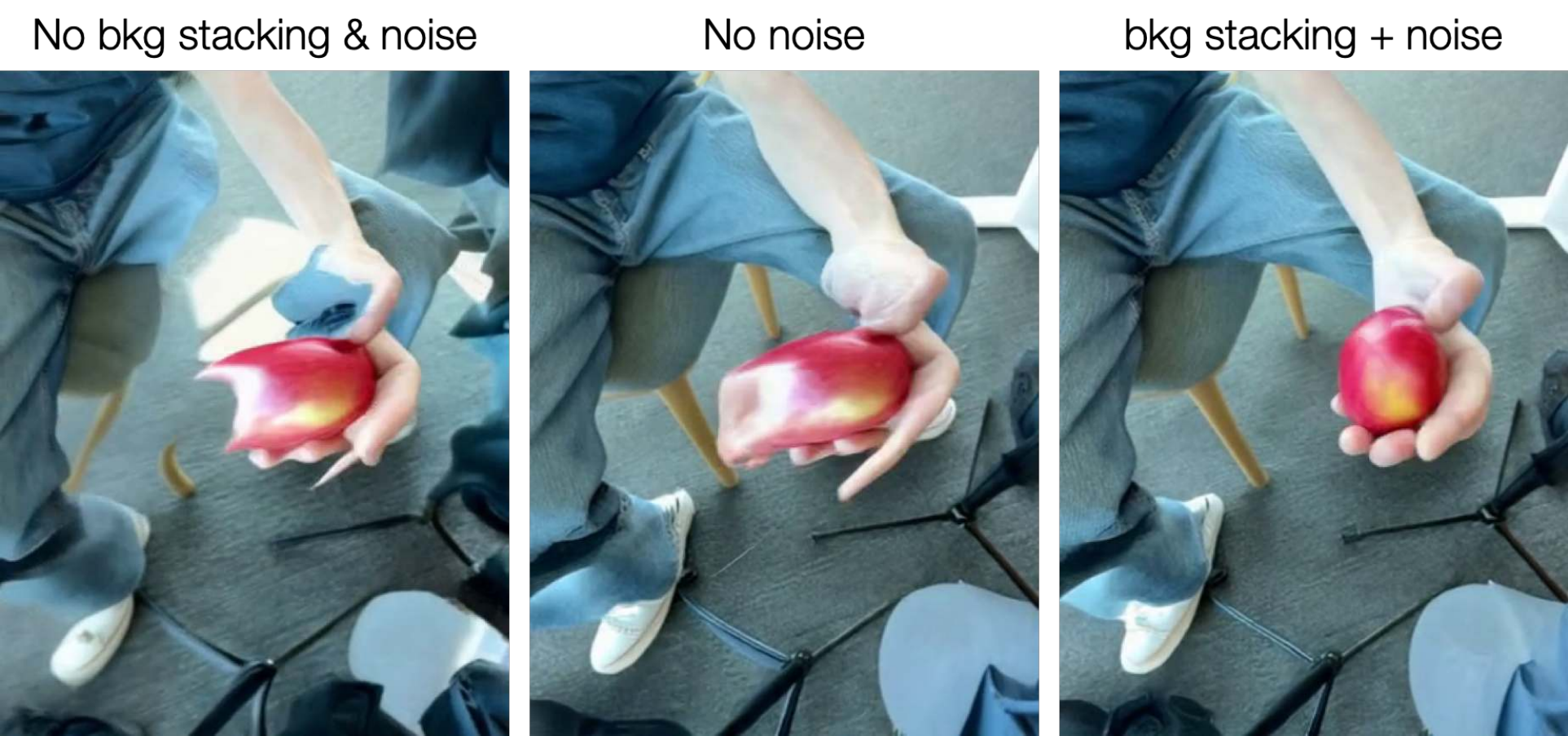}
    \caption{We visualize the `apple' evaluation sequence from DyCheck for analysis of the effect of background stacking over time and noise addition strategy during training to simulate realistic in-the-wild scenarios. First (column 1 vs. 2), we see that for longer videos, stacking the static background from the entire input video helps accumulate multi-view cues about the static background. Second (column 2 vs. 3), we see that due to the noise addition strategy during training, CogNVS is more robust to real-world noise patterns like smearing across object (in this case, apple) edges.}
    \label{fig:ablation_bkg_stack}
\end{figure}

\paragraph{Ablation on background stacking and noise addition} We conduct an ablation on the MegaSAM reconstructions of DyCheck for the effect of static background stacking described in the previous section. In Tab. \ref{tab:bgnoise} and Fig. \ref{fig:ablation_bkg_stack}, we see that stacking the background on DyCheck provides a large in photometric performance. Secondly, we propose to add noise to dynamic object depths during training, especially for out-of-distribution data. This is essential as our creation of self-supervised training pairs only masks out certain pixels from the source video which leaves no room for CogNVS to be able to see real-world noise. To simulate real noise, at say object edges, we estimate the noise between (pseudo) groundtruth depth (coming from iPhone LiDARs or a state-of-the-art depth estimator, say, MoGe) and the predicted depth (coming from a SLAM framework like MegaSAM). This estimated noise for the source pixels, is added to the visible scene reconstruction but in the ray direction of the pixels visible in the arbitrary cameras. This results in noisy visuals that make CogNVS training more robust, especially to out-of-distribution cases. In Tab. \ref{tab:bgnoise} and Fig. \ref{fig:ablation_bkg_stack}, we demonstrate the improvements in performance by training CogNVS to inpaint in the presence of distracting noise artifacts.

\begin{table}[h]
\centering
\caption{We quantitatively evaluate the effect of static background stacking and noise addition on DyCheck. Note that background stacking helps DyCheck because the video sequences are longer than 49-frames that CogNVS can handle. This gives us 3 points performance boost in PSNR. Adding real-world noise to dynamic objects helps make CogNVS robust to noise and therefore it reduces artifacts like smeared object edges, reflected in a much lower FID metric.}
\label{tab:bgnoise}
\resizebox{0.85\columnwidth}{!}{
\begin{tabular}{cccccccccc}
\toprule
Background stacking & Noise addition & PSNR ↑ & SSIM ↑ & LPIPS ↓ & FID ↓ & KID ↓ \\
\midrule
\xmark & \xmark & 11.55 & 0.304 & 0.848 & 197.93 & 0.204 \\
\cmark & \xmark & 14.27 & 0.352 & \textbf{0.737} & 180.67 & 0.171 \\
\cmark & \cmark & \textbf{14.30} & \textbf{0.354} & 0.740 & \textbf{156.83} & \textbf{0.141} \\
\bottomrule
\end{tabular}}
\end{table}

\paragraph{Evaluation with masked metrics on DyCheck} In addition to the metrics reported in the main paper, we also report masked photometric errors as proposed by a prior work \cite{iphone}. While this metric only evaluates the visible scene content and how any view-dependent changes were handled during novel-view synthesis, it does not encourage the generation of unseen scene regions. On this metric, CogNVS performs competitively as compared to baselines.

\begin{figure}[h]
    \centering
    \includegraphics[width=1.0\linewidth]{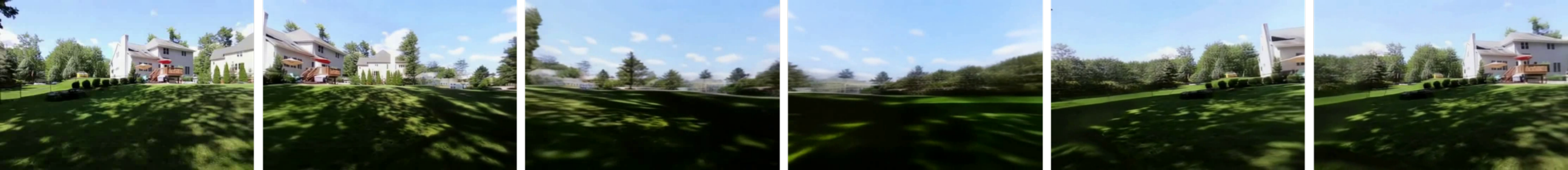}
    \caption{We create a panoramic view of the outdoor scene through progressive generation. Specifically, we divide the 360° camera trajectory into four 90° segments of 49 frames each, reconstructing the previously generated views before reprojecting and inpainting the next 90° novel view. This qualitative analysis suggests that CogNVS can still generate 3D plausible and temporally coherent novel-views even in such extreme cases.}
    \label{fig:360_gen}
\end{figure}

\paragraph{Maximum novel view synthesis angle} As demonstrated in Kubric 4D evaluation and several in-the-wild examples, CogNVS can plausibly generate novel views with up to 90 degrees of variation (up, down, left, or right) in general with a single forward pass. Moreover, our model can produce even more extreme viewpoints by progressively generating new views conditioned on past generations – a fairly straightforward and common strategy adopted by many generative models~\cite{hollein2023text2room, yu2024wonderjourney, 4dim}. We illustrate such a progressively generated panoramic view in Fig.~\ref{fig:360_gen}. Additionally, we want to point out that our method is not able to generate table top views of object-centric scenes (inward facing 360 degree view) beyond 90 degrees of camera deviation, likely because such training data was not seen by the model during pretraining.

\begin{table}[t]
\centering
\caption{We report masked perceptual quality metrics as proposed by prior work \cite{iphone}. This metric only evaluates the visible regions of the scene and so does not encourage generation of unseen scene components. Note that our method performs competitively as compared to the baselines which only focus on modeling the visible scene content.}
\label{tab:sotadycheck}
\resizebox{0.6\columnwidth}{!}{
\begin{tabular}{lccc}
\toprule
Method   & mPSNR ↑ & mSSIM ↑ & mLPIPS ↓ \\
\midrule
MegaSAM \cite{li2024megasam}  & 14.60 & 0.517 & 0.609  \\

Shape-of-Motion \cite{wang2024shapeom}  &  16.47 & \cellcolor{tabfirst}0.639 & \cellcolor{tabsecond}0.409  \\
MoSca \cite{lei2024mosca}    &  \cellcolor{tabfirst}17.82	& \cellcolor{tabsecond}0.635 & \cellcolor{tabthird}0.507 \\
CAT4D \cite{wu2024cat4d}   & \cellcolor{tabsecond}17.39 & \cellcolor{tabthird}0.607 & \cellcolor{tabfirst}0.341\\
TrajCrafter \cite{yu2025trajectorycrafter} & 13.60 & 0.518 & 0.663 \\
\midrule
CogNVS (MegaSAM) & 15.35 & 0.549 & 0.557\\
CogNVS (MoSca) & \cellcolor{tabthird}17.33 & \cellcolor{tabthird}0.607 & 0.530 \\
\bottomrule
\end{tabular}}
\end{table}

\newpage
\section{Qualitative comparison on evaluation datasets}
Please see our webpage for videos in addition to the qualitative visuals below.

\begin{figure}[h]
    \centering
    \includegraphics[width=1.0\linewidth]{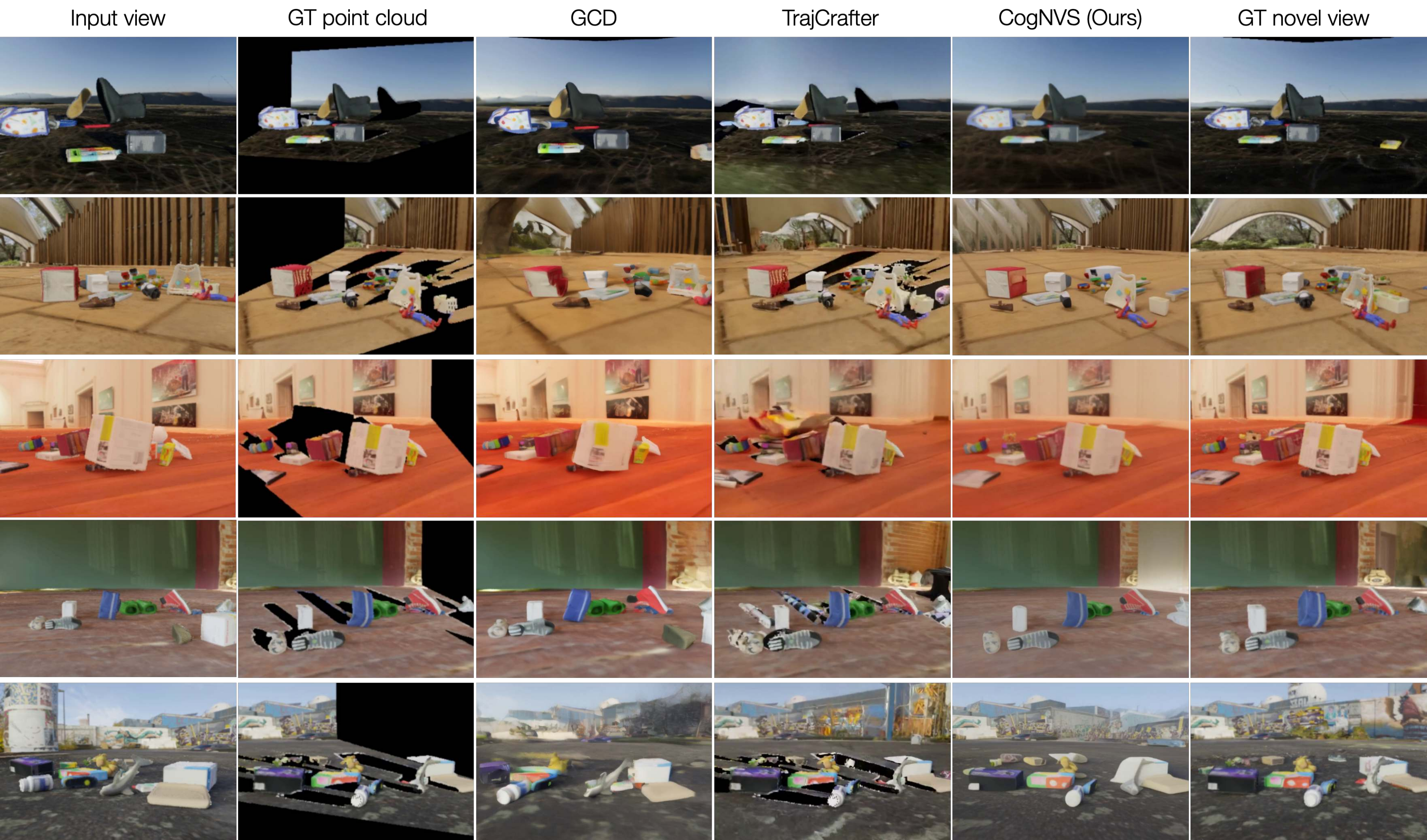}
    \caption{We show supplementary qualitative comparison on Kubric-4D. Note that TrajectoryCrafter is able to generate a reasonable background for the unseen scene regions, but is not able to inpaint the shadows / masks created by foreground objects. GCD is trained on Kubric-4D so performs reasonably well but struggles to preserve the precise geometry. CogNVS achieves better performance as compared to baselines and is the closest is geometric consistency to the groundtruth novel view.}
    \label{fig:comparison_kubric_more}
\end{figure}

\begin{figure}[h]
    \centering
    \includegraphics[width=1.0\linewidth]{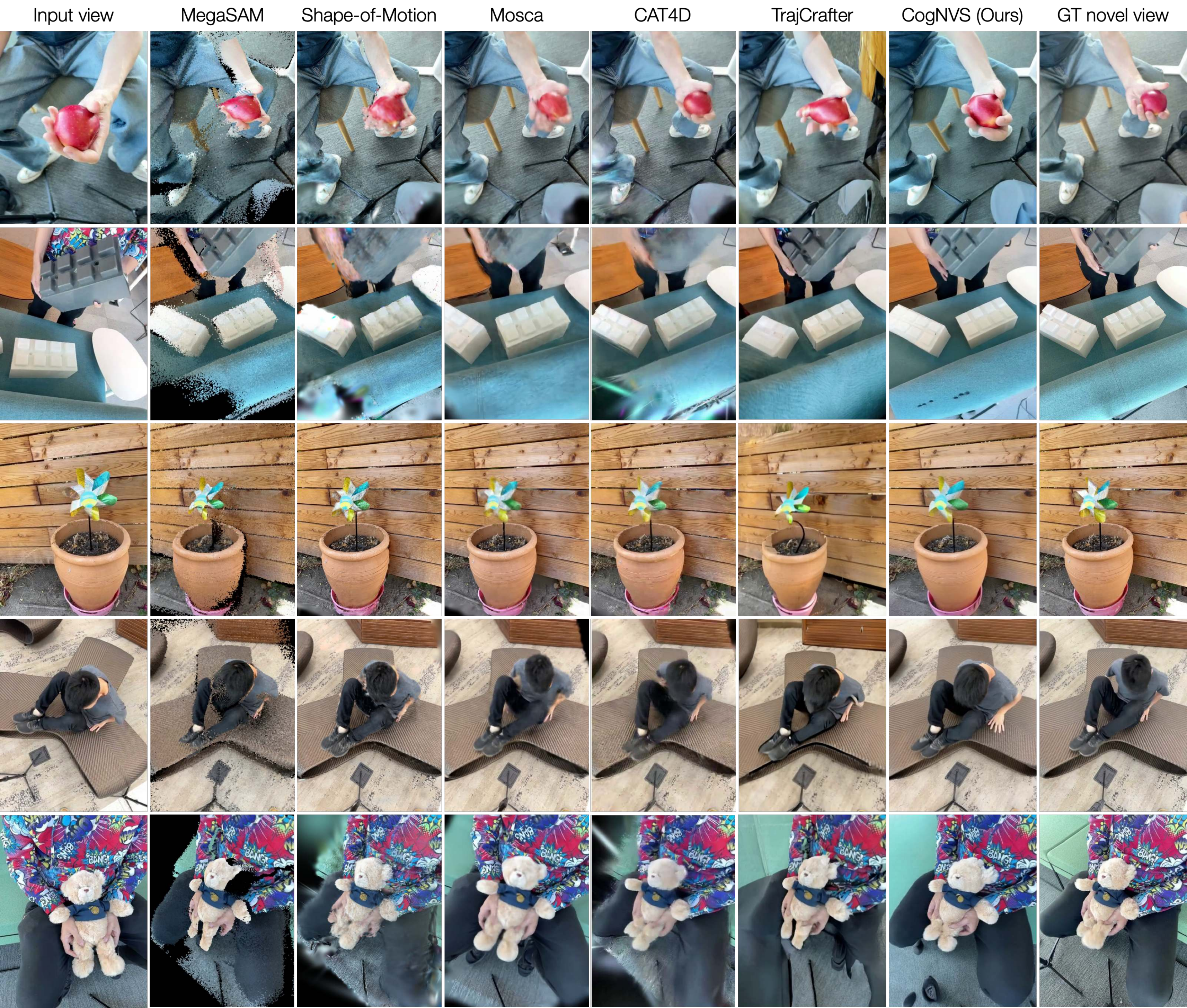}
    \caption{We show supplementary qualitative comparison on DyCheck with CogNVS which surpasses the performance of all prior state-of-the-art. Note that baselines either do not hallucinate the unseen regions in the novel-view (Shape-of-Motion, MegaSAM), show blurry dynamic regions (MoSca, CAT4D), or are not able to preserve the underlying geometry of the scene (TrajectoryCrafter).}
    \label{fig:comparison_dycheck_more}
\end{figure}

\clearpage
\newpage

\section{Qualitative results on in-the-wild examples}

Please see our webpage for videos in addition to the qualitative visuals below.

\begin{figure}[h]
    \centering
    \includegraphics[width=1.0\linewidth]{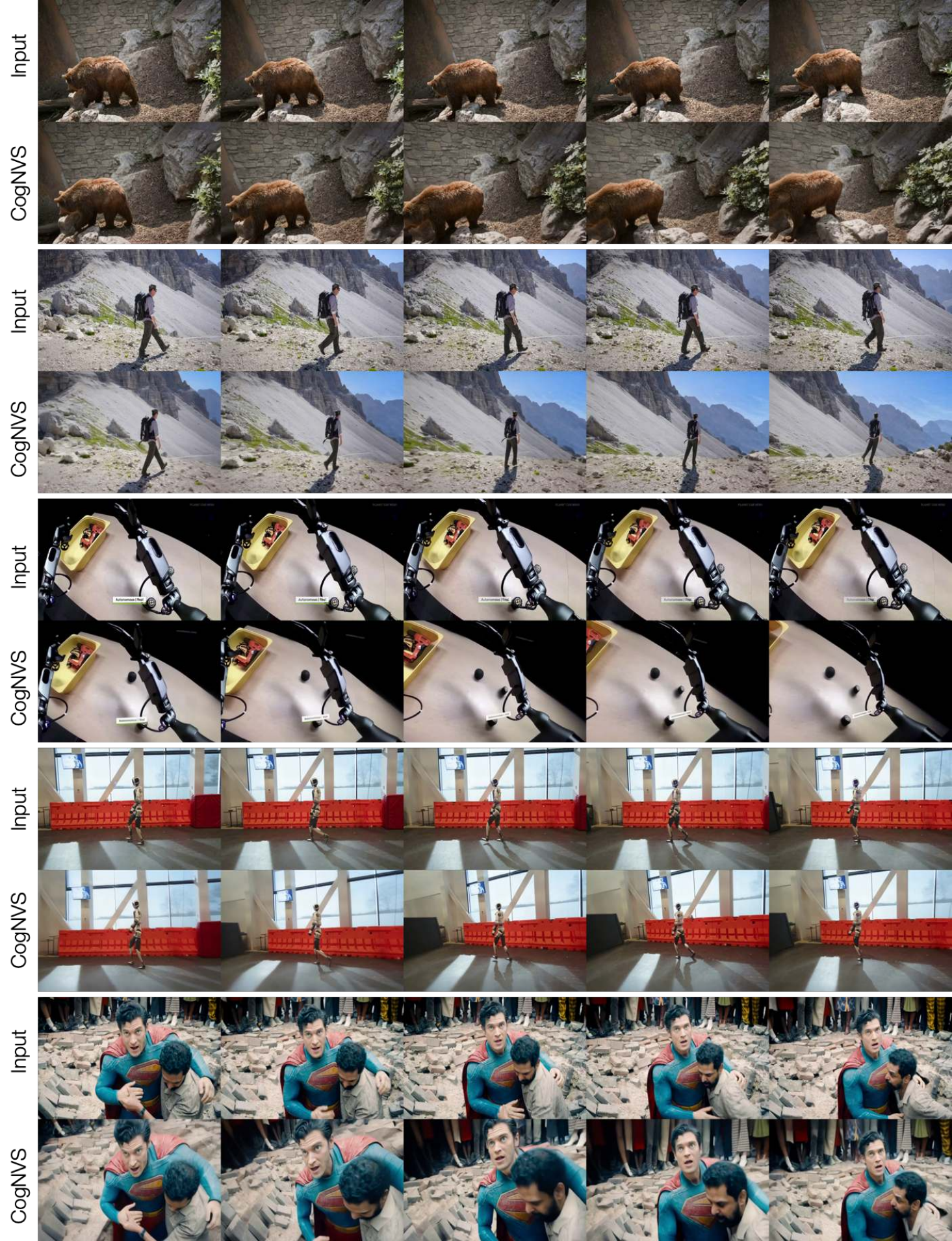}
    \caption{Qualitative results on in-the-wild examples. Part 1 of 2.}
    \label{fig:in_the_wild_real1}
\end{figure}

\begin{figure}[h]
    \centering
    \includegraphics[width=1.0\linewidth]{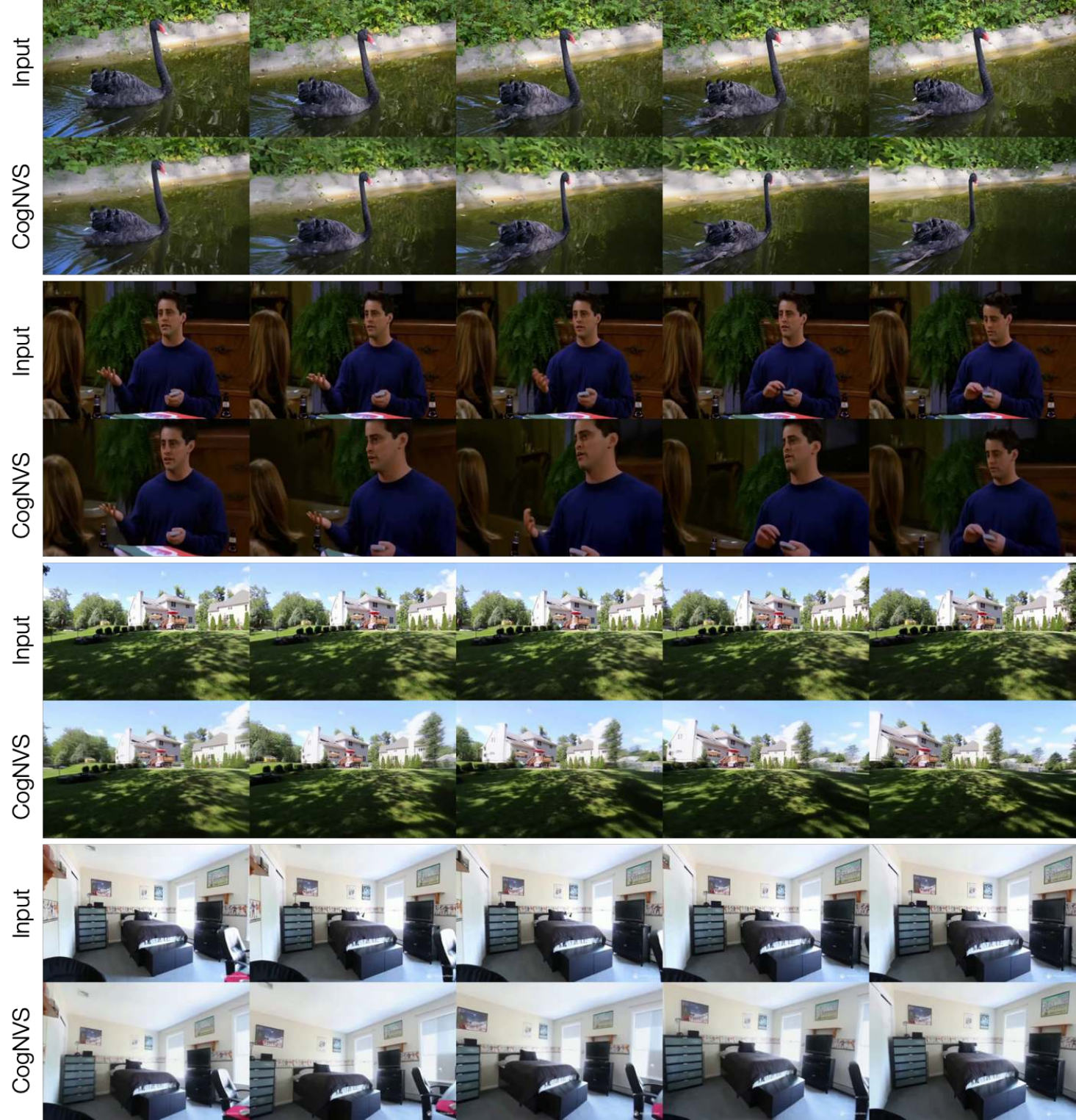}
    \caption{Qualitative results on in-the-wild examples (static scenes included in last two rows). Part 2 of 2.}
    \label{fig:in_the_wild_real2}
\end{figure}

\begin{figure}[h]
    \centering
    \includegraphics[width=1.0\linewidth]{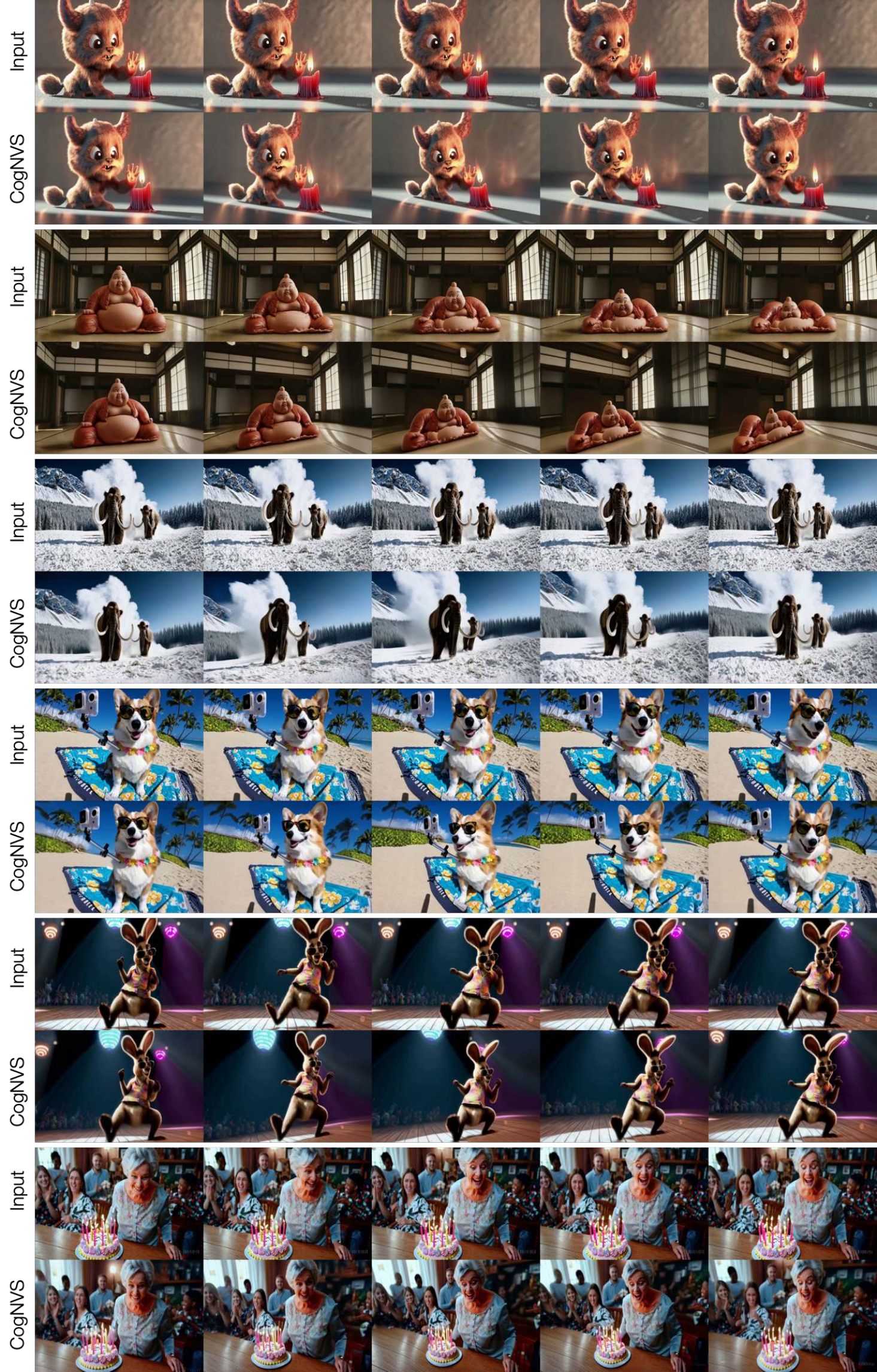}
    \caption{Qualitative results on synthetic videos from SORA.}
    \label{fig:in_the_wild_sora}
\end{figure}

\end{document}